\documentclass[preprint, review,3p,times, 10pt]{elsarticle}

\usepackage{cite}
\usepackage{amsmath,amssymb,amsfonts}
\usepackage{algorithmic}
\usepackage{graphicx}
\usepackage{caption}
\usepackage{subcaption}
\usepackage{textcomp}
\usepackage{comment}
\usepackage{footnote}
\usepackage{comment}
\usepackage{threeparttable}
\usepackage{romannum}
\usepackage{longtable}
\usepackage{lscape}
\usepackage{footnote}
\usepackage{threeparttablex}
\usepackage{multirow}
\usepackage{multicol}
\usepackage{longtable}
\usepackage{soul}
\usepackage{booktabs}
\usepackage{array, makecell}
\usepackage{tikz}
\usepackage{pgfplots}
\usepackage{pgf-pie}
\usepackage{multicol}
\usepackage{multirow}
\usepackage{bbding}
\usepackage[hyphens]{url}
\usepackage{hyperref}
\hypersetup{colorlinks=true,breaklinks=true}

\journal{ArXiv}

\begin{document}

\begin{frontmatter}


\title{ELMF4EggQ: Ensemble Learning with Multimodal Feature Fusion for Non-Destructive Egg Quality Assessment}


\author[1,2]{Md Zahim Hassan}\ead{zahimhassan@baust.edu.bd, zahim951xtgmovi@gmail.com}
\author[1,2]{Md. Osama}\ead{osama_cse@baust.edu.bd, osamasohag39@gmail.com}
\author[3,4]{Muhammad Ashad Kabir \corref{correspondingauthor}}\ead{akabir@csu.edu.au}%
\author[5]{Md. Saiful Islam\corref{correspondingauthor}}\ead{saiful.apma@sau.edu.bd}
\author[5]{Zannatul Naim}\ead{znaim.apma@sau.edu.bd}
\cortext[correspondingauthor]{Corresponding authors.}
\affiliation[1]{organization={Department of Computer Science and Engineering, Bangladesh Army University of Science and Technology}, city={Saidpur}, postcode={5310}, country={Bangladesh}}
\affiliation[2]{organization={Department of Computer Science and Engineering, Rajshahi University of Engineering \& Technology}, state={Rajshahi}, postcode={6204}, country={Bangladesh}}

\affiliation[3]{organization={School of Computing, Mathematics and Engineering, Charles Sturt University}, city={Bathurst}, state={NSW}, postcode={2795}, country={Australia}}
\affiliation[4]{organization={Gulbali Institute for Agriculture, Water and Environment, Charles Sturt University}, city={Wagga Wagga}, state={NSW}, postcode={2678},  country={Australia}}

\affiliation[5]{organization={Department of Animal Production and Management, Sher-e-Bangla Agricultural University}, city={Dhaka}, country={Bangladesh}}

\begin{abstract}
Accurate, non-destructive assessment of egg quality is critical for ensuring food safety, maintaining product standards, and operational efficiency in commercial poultry production. This paper introduces \textit{ELMF4EggQ}, an ensemble learning framework that employs multimodal feature fusion to classify egg grade and freshness using only external attributes -- image, shape, and weight. A novel, publicly available dataset of 186 brown-shelled eggs was constructed, with egg grade and freshness levels determined through laboratory-based expert assessments involving internal quality measurements, such as yolk index and Haugh unit. To the best of our knowledge, this is the first study to apply machine learning methods for internal egg quality assessment using only external, non-invasive features, and the first to release a corresponding labeled dataset.
The proposed framework integrates deep features extracted from external egg images with structural characteristics such as egg shape and weight, enabling a comprehensive representation of each egg. Image feature extraction is performed using top-performing pre-trained CNN models (ResNet152, DenseNet169, and ResNet152V2), followed by principal component analysis (PCA)-based dimensionality reduction, synthetic minority oversampling technique (SMOTE) augmentation, and classification using multiple machine learning algorithms. An ensemble voting mechanism combines predictions from the best-performing classifiers to enhance overall accuracy. Experimental results demonstrate that the multimodal approach significantly outperforms image-only and tabular (shape and weight) only baselines, with the multimodal ensemble approach achieving 86.57\% accuracy in grade classification and 70.83\% in freshness prediction.
The framework demonstrates strong potential for real-time, low-cost deployment in commercial egg processing environments. It highlights the feasibility of using computer vision and lightweight structural inputs for scalable, non-invasive egg quality evaluation.
All code and data are publicly available at \url{https://github.com/Kenshin-Keeps/Egg_Quality_Prediction_ELMF4EggQ}, promoting transparency, reproducibility, and further research in this domain.
\end{abstract}

\end{frontmatter}







\section{Introduction}
\label{sec:introduction}

With the ongoing growth of the global population, there is a steadily increasing demand for food sources that are both affordable and nutritionally rich. Among the available options, eggs have emerged as a particularly important staple due to their high-quality protein content, relatively low cost, and well-balanced nutritional profile~\citep{BRASIL2022108418}. As a widely consumed food item, eggs contribute significantly to human nutrition by providing essential macronutrients and micronutrients, including proteins, fats, vitamins, and minerals. Their biological value, especially in terms of protein quality, is considered one of the highest among natural food sources, comparable to meat, dairy products, and legumes~\citep{sharaf2019egg}. The significance of eggs as a global food source is underscored by the World Egg Organization (WEO), which recognizes the second Friday of October each year as World Egg Day to promote awareness of their nutritional and economic value~\citep{worldeggday}. Eggs are not only an affordable source of high-quality protein and relatively low in calories (about 140 kcal per 100 gm) but also provide a rich supply of essential fats, vitamins, and minerals~\citep{rafed2024nutritional}.

However, nutritional value alone does not ensure consumer acceptance. Consistent quality assurance is critical for both producers and consumers to maintain market standards and ensure food safety. The term ``quality" refers to the level of excellence of a product or its suitability for a specific purpose~\citep{soltani2015egg}. Quality assessment plays a crucial role in ensuring product consistency, extending shelf life, and safeguarding consumer health~\citep{omid2013expert}. To maintain the nutritional and commercial value of eggs throughout the supply chain, effective and reliable quality assessment methods are essential. Egg quality encompasses external characteristics, such as size, and internal attributes, primarily yolk freshness and albumen (egg white) thickness, which degrade over time~\citep{Feddern2017, ABDANANMEHDIZADEH2014105}. External features such as egg size and weight significantly influence consumer perception of quality~\citep{nematinia2018assessment}. Traditionally, egg quality has been assessed using manual techniques such as candling, thermal imaging~\citep{zhang2023nondestructive}, hyperspectral imaging~\citep{dai2020nondestructive}, and destructive methods that require breaking the egg to observe internal attributes~\citep{ab2018automated, abaci2023examination, sehirli2022application, nematinia2018assessment}. While effective, these techniques are labor-intensive, time-consuming, subjective, and often unsuitable for automated systems~\citep{nasiri2020automatic}. As a result, non-destructive testing methods, including shell texture analysis, have gained prominence for their ability to evaluate egg quality rapidly, consistently, and without damaging the product~\citep{nematinia2018assessment}, forming the basis of our study's innovative framework.

Regarding the physical characteristics of an egg, the texture of the shell is a key indicator of quality, reflecting its structural strength, freshness, and ability to resist microbial contamination~\citep{gao2025research, rho2024non, zhang2023nondestructive, liu2020non}. A smooth, uniform texture generally reflects optimal calcification during shell formation, whereas roughness, cracks, or abnormal ridges may signal physiological stress, aging, or nutritional deficiencies in laying hens. Such texture anomalies, including repaired cracks and porous surfaces, can compromise the eggshell’s barrier properties, leading to increased susceptibility to bacterial penetration and moisture loss~\citep{atwa2024advances, poltryegg}. Notably, eggshell translucency and irregular surface patterns have been associated with weakened shell membranes and heightened microbial risk, particularly in older flocks and under fluctuating environmental conditions~\citep{ren2023eggshell}. Comparative studies further reveal that eggshell surface morphology, including roughness and pigmentation, is shaped by ecological and evolutionary factors, reinforcing its biological and functional significance~\citep{attard2023surface}.
While external features, such as shell texture, are critical for consumer appeal and safety, internal quality metrics, such as yolk freshness and albumen thickness, are equally vital for assessing nutritional value. 

Among the significant parameters, the yolk index (YI) is a significant indicator for assessing egg freshness and overall internal quality. The theory behind the low yolk index method is rooted in the physical changes an egg undergoes as it ages. As an egg loses its freshness, its internal quality deteriorates, leading to a loss of structural integrity. This is specifically manifested as the yolk becoming more flattened and spreading out, while the egg white (albumen) also thins and loses viscosity. These changes result in a decrease in the ratio of yolk height to yolk diameter. A lower yolk index value, therefore, indicates an older and less fresh egg~\citep{rafed2024nutritional}. The Haugh unit (HU) is another well-established metric, widely used in commercial and research settings to evaluate the grade of eggs~\citep{Yuan2023}. It is calculated based on the height of the thick albumen relative to the weight of the egg when broken onto a flat surface. Higher HU values indicate better egg quality. As eggs age during storage, the albumen becomes thinner and the HU value decreases, providing a quantitative means of grading egg quality. In this study, we employ the Haugh unit as the reference standard for egg grading, while the yolk index serves as the primary indicator for freshness detection.

Recent advancements in artificial intelligence and computer vision, particularly machine learning and deep learning, have enabled a shift toward fully automated, image-based methods for evaluating egg quality. However, many existing approaches~\citep{dai2020nondestructive, Hsiao2021, OLAKANMI2023111355} focus exclusively on either internal imaging modalities (e.g., hyperspectral or X-ray) or external visual features, without leveraging the correlation between the two. To address the limitations of traditional methods and the disconnect between internal and external quality assessments, this study presents a comprehensive approach that bridges internal and external data sources to improve the accuracy and reliability of egg quality assessment. The main contributions of this work are as follows:
\begin{itemize}
\item We construct and introduce a novel egg dataset with labels for both egg grade and freshness level. This dataset supports supervised learning for quality evaluation tasks and promotes further advancements in the field.

\item We propose an ensemble-based multimodal feature fusion framework that combines egg external image features with egg shape and weight to improve egg grade and freshness classification.

\item We evaluate the effectiveness of using external egg images as a low-cost, non-destructive method for predicting both grade and freshness. This approach is particularly suitable for real-time applications in commercial egg processing facilities.

\end{itemize}

\section{Related Work} 
A diverse body of research has emerged focusing on egg quality measurement. Various studies have proposed different features as indicators of egg quality, including internal quality grading using egg-specific features, external assessments based on conventional or processed images (such as thermal imaging), grading based on egg shape indices, evaluation of volume, weight, and size, detection of defects such as cracks, dark spots or double yolks, and the assessment of egg freshness. 

A substantial body of research has been dedicated to various aspects of egg quality assessment. Many studies have focused on external appearance analysis, including the detection of surface defects such as cracks, dirt, and blood spots~\citep{huang2023damage, turkoglu2021defective, mota2019defect, mota2019defect, guanjun2019cracked, priyadumkol2017crack}. Others have investigated size, weight, and volume estimation to classify eggs into different grades~\citep{okinda2020egg, javadikia2011measuring, thipakorn2017egg, aragua2018cost, siswantoro2017computer, widiasri2019computer}. Internal quality evaluation has also been explored, with methods addressing dark spot detection and double yolk identification~\citep{jiang2021detecting, ma2017identification}. In addition, several works have targeted freshness assessment through non-destructive techniques, including infrared thermal imaging and spectral analysis of eggshell properties~\citep{qin2018research, kim2022egg, sun2015egg, dai2020nondestructive, harnsoongnoen2021grades}. Overall, these studies demonstrate the diversity of features and modalities considered for automated egg quality assessment.

While external physical features such as shell thickness and egg shape provide preliminary information about an egg’s quality, they do not always reliably reflect its internal condition. In practice, internal quality indicators, such as the Haugh Unit and Yolk Index, are more robust measures of egg grading and freshness. The Haugh unit~\citep{jones2012haugh}, derived from the relationship between egg weight and albumen height, is a widely accepted international standard for grading eggs. Similarly, the yolk index serves as a critical parameter for evaluating the deterioration of yolk quality over time. Accurate and non-destructive evaluation of these internal parameters is crucial, especially for meeting consumer expectations and regulatory requirements regarding egg freshness and grading. Thus, methods that can directly estimate internal quality hold particular significance over approaches relying solely on external characteristics.

Table~\ref{tab:literature_review_summary} summarizes several notable works in the field of egg quality assessment. \citet{yang2023computer} developed a system for categorizing eggs based on external defects and predicting weight using major and minor axes, without addressing internal freshness. Similarly, \citet{omid2013expert} proposed an expert system for grading eggs into five quality categories based on surface conditions and size-related features. \citet{dai2020nondestructive} and \citet{zhang2023nondestructive} explored freshness estimation; however, their methods relied on processed image features, such as spectral characteristics or thermal imaging, rather than directly using biological indicators.
To the best of our knowledge, only our proposed method directly integrates internal quality grading based on the Haugh Unit and freshness detection using the Yolk index. By combining multiple feature modalities, including external image, shape index, and weight measurement, our approach uniquely addresses the internal quality and freshness evaluation of eggs within a single, unified framework.

\begin{table}[!ht]
\caption{Summary of related work on egg quality assessment}
\label{tab:literature_review_summary}
{\footnotesize
\begin{tabular}{@{\extracolsep{2pt}}l ccc cccc p{6cm}}

\hline
\multirow{2}{*}{Study} & \multicolumn{3}{c}{Objective} & \multicolumn{4}{c}{Feature Modality} & \multicolumn{1}{c}{\multirow{2}{*}{Methodology}} \\
\cline{2-4} \cline{5-8}
& \makecell[t c]{Internal\\Grading} & Freshness & Others 
& \makecell[t c]{Image\\(External)} & Shape & Weight & Others & \\
\hline\hline

\citep{yang2023computer} & \XSolidBrush & \XSolidBrush & \Checkmark\textsuperscript{a} 
& \Checkmark & \XSolidBrush & \XSolidBrush & \XSolidBrush 
& Feature fusion using image diffraction patterns and hierarchical clustering; classified using RMTDet + RF. \\

\citep{ab2018automated} & \XSolidBrush & \XSolidBrush & \Checkmark\textsuperscript{b} 
& \Checkmark & \Checkmark & \XSolidBrush & \Checkmark 
& PCA and IGR used for feature selection from preprocessed images; classifiers include KNN. \\

\citep{omid2013expert} & \XSolidBrush & \XSolidBrush & \Checkmark\textsuperscript{c} 
& \XSolidBrush & \Checkmark & \XSolidBrush & \Checkmark 
& Color segmentation in HSV space and fuzzy logic-based inference for egg grading. \\

\citep{sehirli2022application} & \XSolidBrush & \XSolidBrush & \Checkmark\textsuperscript{d} 
& \XSolidBrush & \XSolidBrush & \Checkmark & \Checkmark 
& Normalization and 10-fold cross-validation of extracted statistical features; classification with logistic regression. \\

\citep{harnsoongnoen2021grades} & \XSolidBrush & \XSolidBrush & \Checkmark\textsuperscript{e} 
& \Checkmark & \XSolidBrush & \XSolidBrush & \Checkmark 
& Grayscale conversion, background removal, and weight ratio estimation; classification using logistic regression. \\

\citep{dai2020nondestructive} & \XSolidBrush & \XSolidBrush & \Checkmark\textsuperscript{f} 
& \Checkmark & \XSolidBrush & \XSolidBrush & \Checkmark 
& Spectral calibration and noise filtering of hyperspectral images; ensemble method used for classification. \\

\citep{zhang2023nondestructive} & \XSolidBrush & \XSolidBrush & \Checkmark\textsuperscript{g} 
& \XSolidBrush & \XSolidBrush & \XSolidBrush & \Checkmark 
& Thermal video processing with frame extraction, edge detection, and feature selection; classification using SegNet+SVM. \\

\hline
Ours & \Checkmark & \Checkmark & -- 
& \Checkmark & \Checkmark & \Checkmark & -- 
& Ensemble learning with multimodal feature fusion \\
\hline
\multicolumn{9}{l}{\textsuperscript{a} Classifies eggs into five categories (intact, cracked, bloody, floor, and non-standard) and predicts weight using geometric features.}\\
\multicolumn{9}{l}{\textsuperscript{b} Grades eggs (AA, A, B, C, D, E) based on egg shape parameters.}\\
\multicolumn{9}{l}{\textsuperscript{c} Grades eggs (Excellent, Good, Medium, Bad, Wastage) using characteristics such as cracks, blood spots, breakage, and size.}\\
\multicolumn{9}{l}{\textsuperscript{d} Uses 20 features for binary classification into acceptable or unacceptable quality.}\\
\multicolumn{9}{l}{\textsuperscript{e} Estimates freshness using egg density based on height and width.}\\
\multicolumn{9}{l}{\textsuperscript{f} Applies Haugh unit–based criteria for shell quality assessment, but not yolk index.}\\
\multicolumn{9}{l}{\textsuperscript{g} Proposes a non-destructive method for freshness estimation using thermal imaging.}\\
\end{tabular}
}
\end{table}

Building upon these insights, our work proposes a novel system that takes advantage of easily available egg characteristics to achieve both internal quality grading and freshness classification. Unlike prior studies that often focused on isolated aspects of quality or employed complex and expensive imaging techniques, our approach emphasizes practical applicability by utilizing standard features such as external images, shape indices, and weight measurements.

\section{Dataset}

\subsection{Egg Collection}

A total of 288 brown-shelled chicken eggs were collected from Dhaka, Bangladesh. To capture variability in origin, handling practices, and storage conditions, eggs were sourced equally (n = 72) from four distinct market categories: wholesale market (WM), super shop (SS), grocery shop (GS), and open shop (OS). After collection, eggs were transported to the laboratory on the same day of collection to preserve freshness and quality. The eggs originated from a diverse range of farms, sources both within and outside Dhaka. Each market category reflects a unique distribution channel. Wholesale markets (WM) serve as major distributors supplying large volumes of eggs to other retailers. Super shops (SS) obtain eggs either from wholesale markets or through corporate supply chains. Grocery shops (GS) typically sell eggs alongside various daily necessities, while open shops (OS) operate through small independent stalls or mobile vendors that sell eggs directly to consumers. 

This stratified and balanced sampling design was implemented to capture real-world market variability in egg supply chains. By including diverse market types, the dataset reflects the variation in sourcing, storage, and handling practices that consumers are exposed to. This diversity is crucial for evaluating how different retail conditions influence the external and internal quality of eggs, as well as their safety for consumption. All eggs were handled under controlled laboratory conditions upon arrival to maintain the integrity of quality.



\subsection{Dataset Preparation}
Following collection from the four distinct market categories, all eggs were stored under standard refrigeration conditions to ensure consistency in post-collection handling. At regular intervals of seven days, a random subset of eggs from each market group was selected for data acquisition. This process involved capturing external images and conducting internal assessments by breaking the eggs. 
This staggered sampling protocol was designed to introduce variability in egg freshness and quality, thereby allowing the dataset to reflect a broader spectrum of egg grades over time. By simulating real-world storage durations, the resulting dataset incorporates natural degradation patterns and quality transitions, which are critical for training and evaluating models aimed at freshness prediction and quality classification.

\begin{figure}[b]
    \centering
    \includegraphics[width=0.6\linewidth]{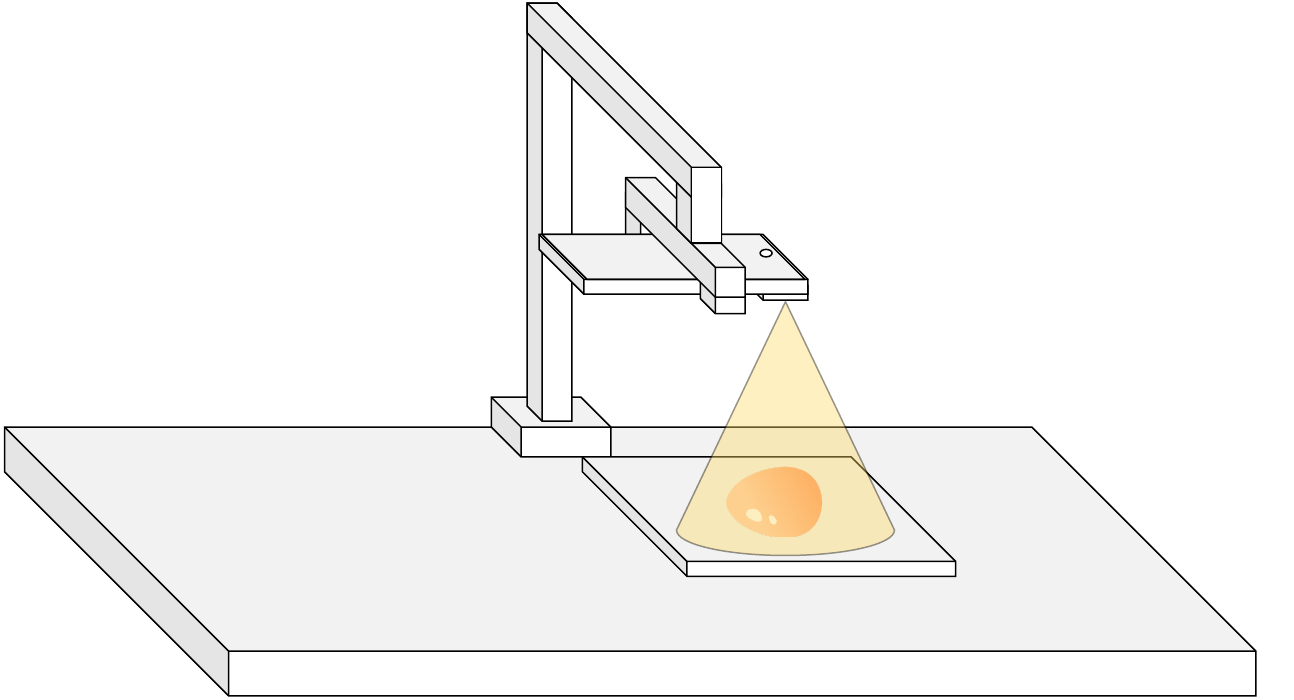}
    \caption{Image capturing setup for individual egg sample.}
    \label{fig:image_capture}
\end{figure}

Each egg underwent standardized image capturing to support downstream analysis. All illustrated in Figure~\ref{fig:image_capture}, all images were taken using a fixed digital camera setup positioned 15 cm above the egg surface, under controlled and uniform lighting conditions. This imaging process ensured visual uniformity across the dataset and supports future applications in computer vision-based egg quality assessment. Following image capture, the eggs were promptly taken to the controlled environment to preserve freshness and internal quality.
To evaluate the physical characteristics of the collected eggs, two primary external attributes were measured: egg weight and shape index. 
Egg weight was measured using a precision digital weighing scale with an accuracy of $\pm0.01$ grams, with consistent procedures~\citep{Sapkota2020}. It plays a central role in egg classification and quality grading, particularly in retail and food safety contexts.

\begin{figure}[!ht]
\centering
\begin{subfigure}{.31\textwidth}
  \centering
  \includegraphics[width=\textwidth]{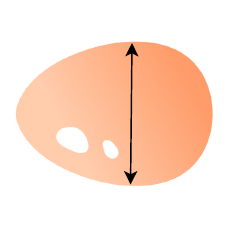}
  \caption{Egg width}
  \label{fig:grade_data}
\end{subfigure}
\hspace{3em}
\begin{subfigure}{.33\textwidth}
  \centering
  \includegraphics[width=\textwidth]{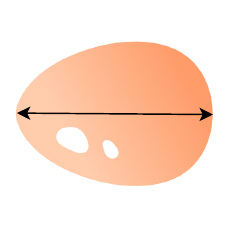}
  \caption{Egg length}
  \label{fig:freshness_data}
\end{subfigure}
\caption{Measurement of Egg Shape Index using width and length.}
\label{fig:shape_index}
\end{figure}

Shape index (SI) was used to describe the proportional relationship between the egg's width and length, offering insights into egg morphology and consumer preference. The shape index was calculated according to the formula~\citep{reddy1981egg} presented in Equation~\ref{eq-5}. 
\begin{equation}
\label{eq-5}
    \text{Shape index (SI)} = \left(\frac{Width}{Length}\right) \times 100
\end{equation}
where $Width$ represents the maximum breadth (width) of the egg and $Length$ is the maximum longitudinal length, illustrated in Figure~\ref{fig:shape_index}. Higher shape index values indicate more spherical eggs.

\begin{figure}[!ht]
\centering
\begin{subfigure}{.26\textwidth}
  \centering
  \includegraphics[width=\textwidth]{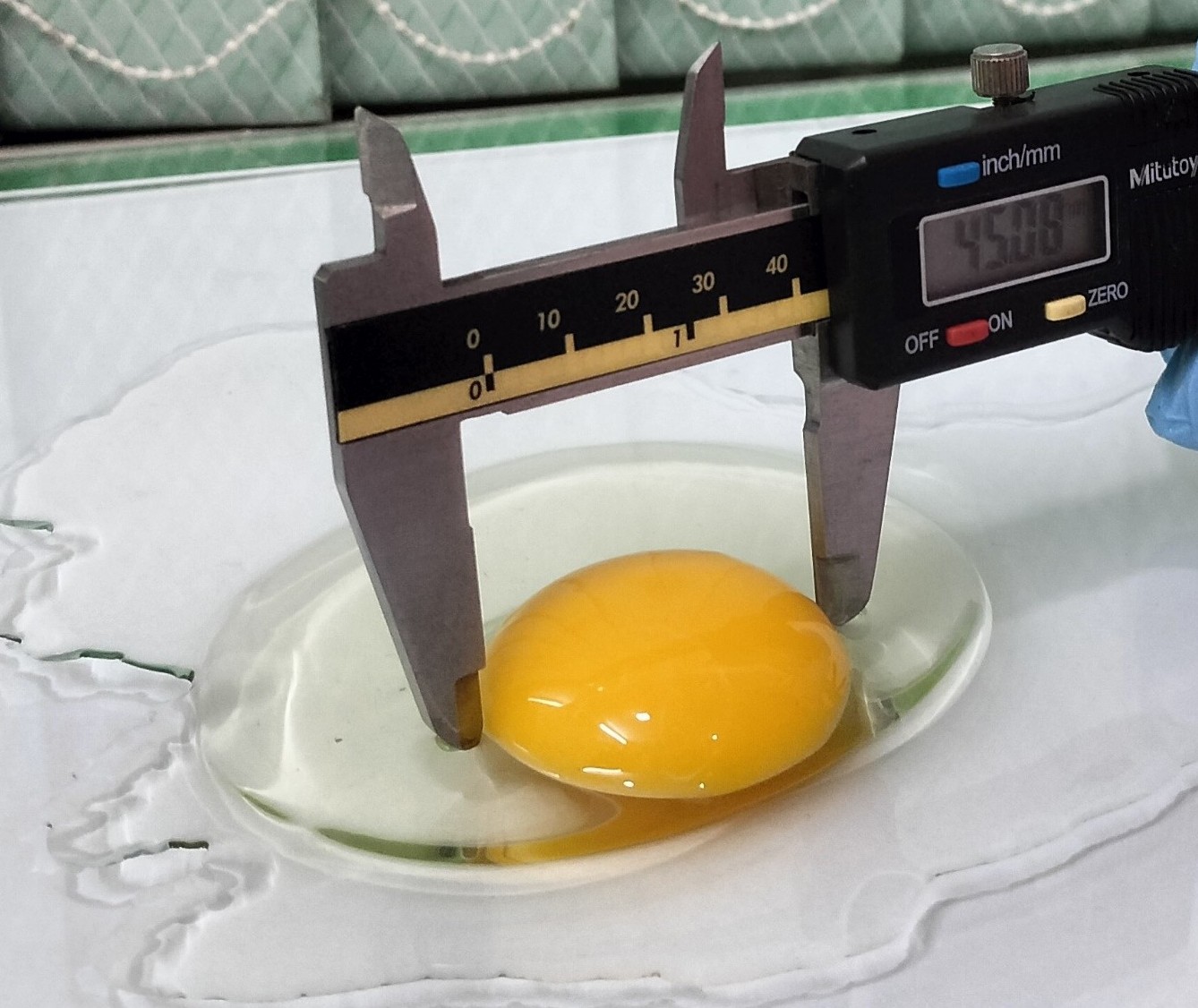}
  \caption{Yolk diameter}
  \label{fig:yolk_diameter}
\end{subfigure}
\hspace{3em}
\begin{subfigure}{.26\textwidth}
  \centering
  \includegraphics[width=\textwidth]{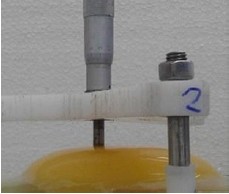}
  \caption{Yolk height}
  \label{fig:yolk_height}
\end{subfigure}
\hspace{3em}
\begin{subfigure}{.26\textwidth}
  \centering
  \includegraphics[width=\textwidth]{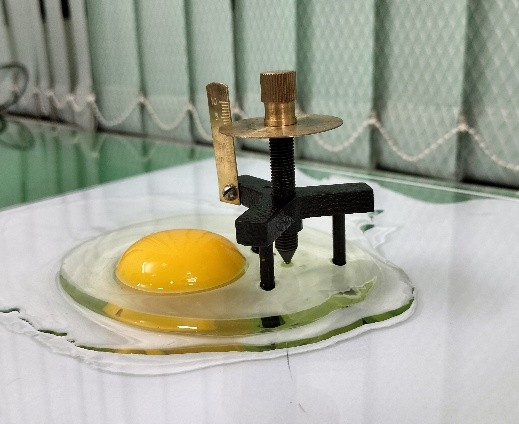}
  \caption{Albumen height}
  \label{fig:albumen_height}
\end{subfigure}
\caption{Measurement of an egg's internal attributes.}
\label{fig:measurement}
\end{figure}

Among internal quality parameters, the yolk index (YI) is widely recognized as a reliable indicator of egg freshness. Yolk diameter was measured using a digital caliper shown in Figure~\ref{fig:yolk_diameter}, while yolk height was determined with a trivet micrometer as shown in Figure~\ref{fig:yolk_height}. The yolk index was then computed using Equation~\ref{yolk}, which expresses the ratio of yolk height to yolk diameter:
\begin{equation}
\text{Yolk index (YI)} = \left(\frac{\text{Yolk height}}{\text{Yolk diameter}}\right) \times 100
\label{yolk}
\end{equation} 

Table~\ref{tab:yolk_class} provides clear thresholds for categorizing eggs into different freshness levels based on their yolk index (YI)~\citep{Huang2012}. Eggs with a YI greater than 38 are considered fresh, reflecting a rounder, firmer yolk and superior internal quality. Eggs with a YI between 34.5 and 38 are classified as moderately fresh, indicating slightly reduced yolk firmness, while values below 34.5 correspond to old eggs with a flatter yolk and diminished quality. This classification system allows for an objective assessment of egg freshness, providing a standardized approach that is useful for both research studies and industrial quality control processes.
\begin{table}[!ht]
    \centering
    \caption{Yolk Index Classification}
    \begin{tabular}{lc}
        \hline
        Freshness category & Yolk index (YI) Range \\
        \hline
        Fresh            & $\text{YI}> 38$ \\
        Moderately fresh & $34.5 \leq \text{YI} \leq 38$ \\
        Old              & $\text{YI}< 34.5$ \\
        \hline
    \end{tabular}
    \label{tab:yolk_class}
\end{table}

The haugh unit (HU) is a widely accepted quantitative measure of internal egg quality. It incorporates both the egg's weight and the height of its thick albumen. HU values were calculated using the formula~\citep{Huang2012} that is given in Equation~\ref{HU} :
\begin{equation}
    \text{HU} = 100 \times \log_{10}\left(H + 7.6 - 1.7 \times W^{0.37} \right)
    \label{HU}
\end{equation}
where $H$ denotes the albumen height, which is measured as depicted in Figure~\ref{fig:albumen_height}, and $W$ is the egg weight in grams. Based on the resulting HU scores, each egg was labeled into quality grades ($AA$, $A$, $B$, or $C$)~\citep{JiangOvomucin}.

Table~\ref{tab:haugh_class} summarizes egg quality grades based on the haugh unit, which incorporates both albumen height and egg weight. Grade $AA$ represents the highest quality eggs with firm albumen, while Grade $C$ denotes eggs of low internal quality. Grades $A$ and $B$ indicate intermediate quality levels based on albumen firmness and egg weight. Together with the yolk index, HU provides a comprehensive framework for internal egg quality grading. These measurements are essential for accurately assessing egg freshness and consistency and support reliable monitoring in quality control systems.
\begin{table}[!ht]
    \centering
    \caption{Haugh unit-based grading}
    \begin{tabular}{cc}
        \hline
        Grade & Haugh unit (HU) range \\
        \hline
        AA & $\text{HU} \geq 72$ \\
        A  & $60 \leq \text{HU} < 72$ \\
        B  & $31 \leq \text{HU} < 60$ \\
        C  & $\text{HU} < 31$ \\
        \hline
    \end{tabular}
    \label{tab:haugh_class}
\end{table}

\subsection{Dataset Description}

During the data collection process, a number of eggs were excluded due to various quality and integrity issues. Specifically, some eggs were accidentally broken during handling, found to be rotten upon inspection, or had damaged yolks that made internal assessment unreliable. In addition, samples with missing measurements, data inconsistencies, or extreme outlier values were systematically identified and removed during preprocessing. After applying these quality control steps, a total of 186 valid samples remained from the initial 288 collected eggs. These cleaned and verified samples form the final dataset used for analysis and model development in this study.

Categorical simplification was applied to ensure a more balanced dataset, as there was an imbalance in the number of samples within the initial classes. The original egg grading scheme, which included four distinct levels ($AA$, $A$, $B$, and $C$), was restructured into two broader categories: samples classified as $AA$ or $A$ were grouped under the ``High" grade class, while those falling into grades $B$ or $C$ were assigned to the ``Low" grade class. Similar to grading, the initial freshness classification, which distinguished among $Fresh$, \textit{Moderately fresh} and $Old$ eggs, was consolidated into two categories. Eggs labeled as either $Fresh$ or \textit{Moderately fresh} were grouped under the unified ``Fresh" class, whereas eggs originally labeled as $Old$ were retained in a separate ``Old" category. This binary categorization improved class balance and enabled more robust model training and evaluation. Table~\ref{tab:egg_stats} presents a summary of the descriptive statistics for the 186 finalized samples. The features include egg weight, shape index, yolk index, and Haugh unit, categorized by the newly defined binary grade (High, Low) and freshness labels (Fresh, Old). Statistics are reported in terms of minimum, maximum, mean, and standard deviation (SD).

\begin{table}[!ht]
\centering
\caption{Descriptive Statistics of Egg Features by Grade and Freshness (n=186)}
\label{tab:egg_stats}
{\small
\setlength{\tabcolsep}{14pt} 
\renewcommand{\arraystretch}{0.8} 
\begin{tabular}{@{\extracolsep{4pt}}llcccc}
\hline
\multirow{2}{*}{Attribute} & \multirow{2}{*}{Measure} 
& \multicolumn{2}{c}{Grade (n=186)} 
& \multicolumn{2}{c}{Freshness (n=186)} \\
\cline{3-4}\cline{5-6}
 & & High (n=78) & Low (n=108) & Fresh (n=90) & Old (n=96) \\
\hline\hline
\multirow{3}{*}{Weight (g)} 
 & Min       & 47.69 & 44.78 & 47.69 & 44.78 \\
 & Max       & 77.15 & 70.30 & 77.15 & 70.30 \\
 & Mean (SD) & 60.13 (4.69) & 58.92 (4.57) & 60.35 (4.62) & 58.57 (4.53) \\
\hline
\multirow{3}{*}{Shape index} 
 & Min       & 69.54 & 71.91 & 69.54 & 71.91 \\
 & Max       & 96.00 & 96.21 & 96.00 & 96.21 \\
 & Mean (SD) & 78.84 (3.82) & 77.47 (3.15) & 78.50 (3.71) & 77.62 (3.27) \\
\hline
\multirow{3}{*}{Yolk index} 
 & Min       & 34.79 & 1.91 & 34.54 & 1.91 \\
 & Max       & 50.12 & 37.25 & 50.12 & 34.33 \\
 & Mean (SD) & 38.66 (2.57) & 25.96 (7.80) & 38.27 (2.62) & 24.74 (7.41) \\
\hline
\multirow{3}{*}{Haugh unit} 
 & Min       & 61.10 & 1.80 & 32.91 & 1.80 \\
 & Max       & 104.50 & 59.70 & 104.50 & 59.70 \\
 & Mean (SD) & 77.91 (9.18) & 41.35 (13.96) & 73.99 (13.52) & 40.45 (14.28) \\
\hline
\multirow{4}{*}{Market (count)} 
 & GS & 17 & 28 & 22 & 23 \\
 & OS & 27 & 26 & 28 & 25 \\
 & SS & 11 & 33 & 13 & 31 \\
 & WM & 23 & 21 & 27 & 17 \\
\hline
\end{tabular}
}
\end{table}

As seen in the table, High-grade and Fresh eggs exhibited better values across multiple features. The yolk index and haugh unit in particular showed strong contrast between High vs. Low grade, and Fresh vs. Old groups, reinforcing their effectiveness as egg quality and freshness indicators. The distribution across market categories remained well represented, enabling reliable comparative and predictive analyses in subsequent modeling tasks.

\section{Methodology}
\begin{figure}[!htp]
    \centering
    \includegraphics[width=1\textwidth]{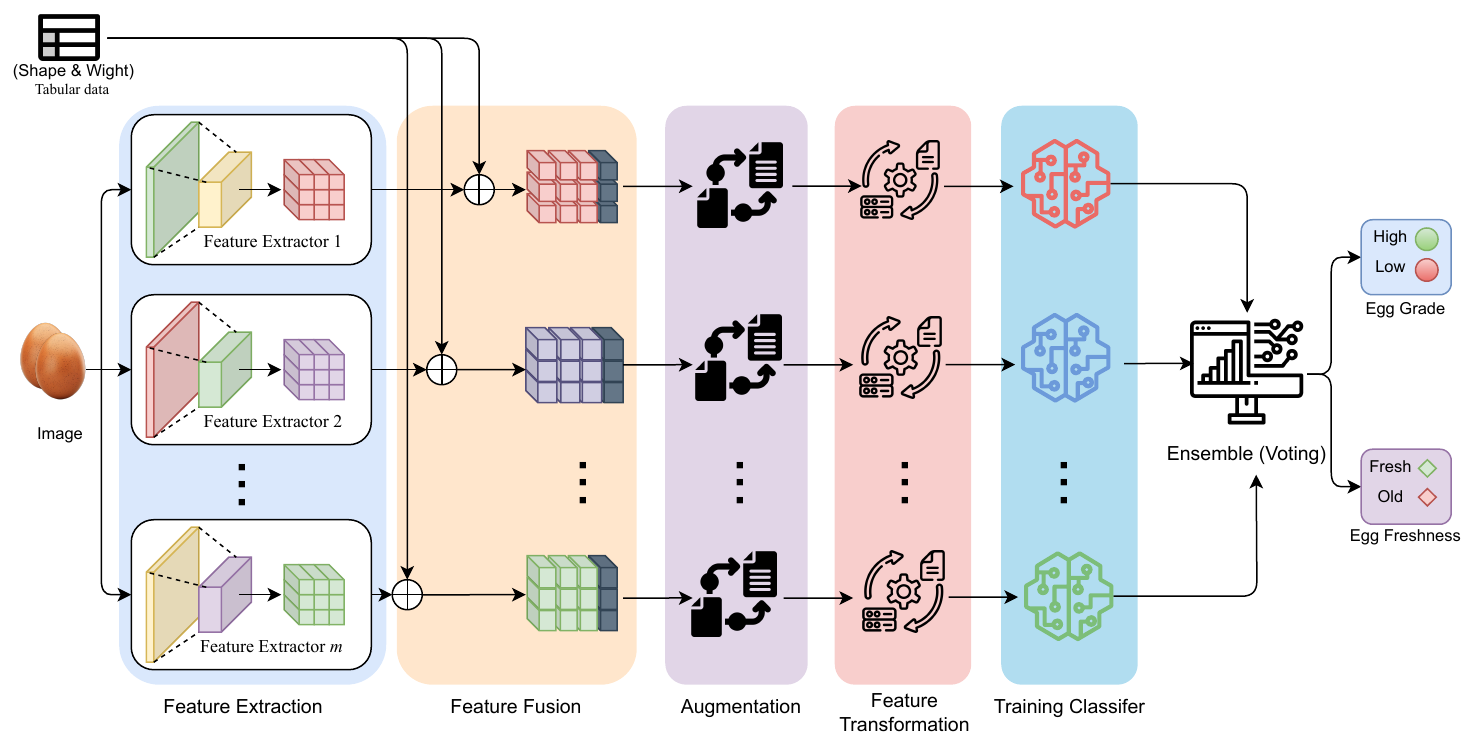}
    \caption{Proposed ensemble-based multimodal feature fusion framework for egg grading and freshness classification}
    \label{fig:method}
\end{figure}

Figure~\ref{fig:method} illustrates the proposed ensemble-based multimodal feature fusion framework designed for automated egg grading and freshness classification. The framework leverages multimodal feature fusion by combining visual features extracted from egg images with egg structural attributes -- specifically shape and weight -- to capture comprehensive information about each egg. It employs multiple feature extractors to derive diverse feature representations from the egg image, which are subsequently fused and passed through a series of modules for augmentation, transformation, and classification. The ensemble mechanism at the final stage leverages majority voting across multiple classifiers to produce robust predictions for both egg grade (high vs. low) and freshness (fresh vs. old).

\subsection{Feature Extraction and Fusion}
Feature extraction is often used in image processing for simplifying image representation, thus enhancing computational efficiency  and improving performance~\citep{mutlag2020feature}. Images contain various types of features, including statistical, color, and texture features, among others. However, extracting each feature type individually from every image is computationally intensive and inefficient. 

To overcome these limitations, especially in scenarios with limited data resources, this study employs transfer learning using pre-trained convolutional neural networks (CNNs) for image feature extraction~\citep{dulal2025mhaff, poola2023covid, chowdhury2021few}. In such settings, models pre-trained on large-scale datasets (e.g., ImageNet) are leveraged by removing their final classification layers, thereby utilizing the learned representations as feature extractors. When an input image is passed through such a model, the output of the final global average pooling layer serves as a high-level feature vector representing the image. We evaluate the effectiveness of multiple state-of-the-art pre-trained CNN architectures for this purpose, including InceptionResNetV2~\citep{szegedy2017inception}, Xception~\citep{chollet2017xception}, ResNet101, ResNet152, ResNet12V2~\citep{he2016deep}, MobileNetV2~\citep{sandler2018mobilenetv2}, DenseNet169, DenseNet201~\citep{huang2017densely}, InceptionV3~\citep{szegedy2016rethinking}, EfficientNetB7~\citep{tan2019efficientnet}, ConvNeXtTiny, ConvNeXtLarge~\citep{liu2022convnet} pre-trained models as the feature extractor. 

Prior to feature extraction, all images were normalized and resized to a resolution of $224\times224$ pixels. For each CNN model, the final prediction layer was excluded, and the output of the last global average pooling layer was retained as the image feature vector. For instance, DenseNet169 yields a feature matrix of size $n \times 1664$ for $n$ input images, while ResNet152 and ResNet152V2 produce matrices of size $n \times 2048$. 


In addition to image-based features, tabular data comprising structural attributes such as egg shape and weight were also incorporated. A simple sample-wise concatenation strategy was employed to fuse these numerical features with the extracted image features, resulting in a comprehensive multimodal feature representation for each egg.

\subsection{Augmentation}
\label{preprocessing_augmentation}

To address the issue of class imbalance in the dataset, an augmentation technique called Synthetic Minority Oversampling Technique (SMOTE) is used to increase the samples to eliminate the class imbalance problem~\citep{chawla2002smote}. This method involves selecting an instance from the minority class and identifying its nearest neighbors. A new instance is generated by interpolating within the feature space between the original instance and its neighbors. Through this process, SMOTE produces additional instances of the minority class that closely resemble existing samples, thus increasing the sample size of the minority class and mitigating class imbalance. This method increases the number of minority class samples in a way that preserves the underlying data distribution while avoiding exact duplication of existing instances. As a result, SMOTE enhances the model's ability to generalize across underrepresented classes and mitigates the risk of biased learning due to class imbalance.

In this study, SMOTE was applied to the merged multimodal feature set comprising both image-derived and tabular features to generate a balanced dataset across all target classes.
Table~\ref{tab:before_after_data_distribution} reports the class distributions before and after the SMOTE operation, highlighting that after performing augmentation, the dataset becomes perfectly balanced. 


\begin{table}[!ht]
\centering
\caption{Sample distribution of target classes (before and after augmentation)}
\label{tab:before_after_data_distribution}
\begin{tabular}{llcc}
\toprule
Task              & Category & Before Augmentation & After Augmentation \\
\hline
\hline
\multirow{2}{*}{Egg Grade Prediction}   & High              & 78                           & 108                         \\
                           & Low               & 108                          & 108                         \\
\hline
\multirow{2}{*}{Egg Freshness Prediction} & Fresh             & 90                           & 96                          \\
                           & Old               & 96                           & 96                          \\
\hline
\end{tabular}
\end{table}


\subsection{Feature Transformation}
\label{section:PCA}
High-dimensional feature spaces often introduce computational overhead, particularly in real-time applications, and may include redundant or non-informative variables that do not significantly contribute to the predictive performance of the model. To address these challenges, this study applies Principal Component Analysis (PCA)~\citep{mackiewicz1993principal} as a dimensionality reduction technique aimed at transforming the feature space while preserving 99\% of the original data variance.

This transformation reduces the dimensionality of the feature vectors, resulting in more compact representations that retain the essential patterns and information necessary for classification tasks. The method begins by computing the covariance matrix of the original features to understand the relationships between variables. The covariance between any two features, $w$ and $q$, is calculated using Equation~\ref{eq-2}:
\begin{equation}
    cov(w, q) = \frac{\sum_{i=0}^n (w_i - \mu_w) (q_i - \mu_q)}{n-1}
    \label{eq-2}
\end{equation}
where $n$ represents the number of samples, and $\mu_w$ and $\mu_q$ denote the mean values of features $w$ and $q$, respectively.

Once the covariance matrix is obtained, PCA identifies new orthogonal axes, known as principal components, that maximize the variance in the data. These components are derived from the eigenvectors and eigenvalues of the covariance matrix, with eigenvalues indicating the amount of variance captured by each principal component. The components are ranked in descending order based on the variance they explain, and the top-$k$ components that cumulatively account for at least 99\% of the total variance are retained.
The original high-dimensional data is then projected onto this reduced feature space, resulting in a transformed dataset that is computationally more efficient and less prone to overfitting.


\subsection{Training Classifier}
The transformed feature sets obtained after dimensionality reduction were used to train a variety of machine learning classifiers to evaluate performance across diverse algorithmic paradigms. By incorporating a diverse set of classifiers, this study aims to comprehensively evaluate the effectiveness of the proposed feature sets in classifying egg quality and freshness. The classifiers include: Regression, Decision Tree, Random Forest, Support Vector Classifier (SVC), Gradient Boosting, Multi-Layer Perceptron (MLP), XGBoost, LightGBM, and Adaptive Boosting. Logistic Regression is a statistical method that utilizes the sigmoid function for classification tasks. Decision Tree adopts a decision-based approach, which is to recursively partition a dataset into smaller subsets based on feature values, creating a tree-like structure where each node represents a decision or outcome to classify instances into target classes. 

Random Forest, Gradient Boosting, XGBoost, and LightGBM are ensemble-based models. Random Forest leverages an ensemble of decision trees, each trained on random subsets of data and features, to generate predictions. Gradient Boosting, another ensemble-based technique, constructs decision trees sequentially, with each tree designed to correct the errors of its predecessors. XGBoost uses an extreme gradient boosting algorithm with an ensemble approach for categorization tasks. Similarly, LightGBM employs a light-gradient boosting algorithm, focusing on minimizing the loss function through gradient descent. All gradient boosting methods operate by combining weak learners to form a robust predictive model. Support Vector Classifier (SVC) identifies an optimal hyperplane to separate the data so that prediction can be done based on this boundary. The Multi-Layer Perceptron (MLP), a type of multilayer feedforward neural network, is designed to capture complex, non-linear relationships for classification tasks. Adaptive Boosting, also an ensemble-based method, aggregates the outputs of multiple simpler algorithms to produce a final prediction. 



\subsection{Ensemble}
In the final stage of the proposed framework, predictions from multiple classifiers are aggregated to produce a more robust and accurate final prediction. To achieve this, we employed the majority voting technique, one of the most widely used ensemble methods for classification tasks. In this approach, each individual classifier contributes a vote for the predicted class label, and the class that receives the majority of votes is selected as the final prediction for a given input.
The ensemble decision mechanism can be formally expressed as:
\begin{equation}
    \hat{y} = mode[C_1(x), C_2(x),...C_m(x)]   
    \label{eq-3}
\end{equation}
where $\hat{y}$ denotes the final ensemble prediction, $x$ represents the input sample, and $C_m(x)$ is the predicted label produced by the $m^{th}$ classifier.

This majority voting strategy enhances the generalization capability of the model by leveraging the complementary strengths of individual classifiers and mitigating the impact of any single model's misclassification. By aggregating diverse decision boundaries, the ensemble model achieves improved stability and predictive performance compared to any single classifier in isolation.

\section{Experimental Evaluation}

\subsection{Settings}
To rigorously evaluate model performance for both egg grading and freshness classification tasks, a 10-fold cross-validation strategy was employed. This method ensures a robust and unbiased estimate of each model's generalization capability by partitioning the dataset into 10 equal subsets. In each fold, 9 subsets were used for training and 1 for testing, with the process repeated across all folds. The average classification accuracy across the 10 folds was computed and reported as the final performance metric for each experiment.

To ensure fair comparisons and optimal performance, Grid Search was applied during the cross-validation process to identify the best-performing hyperparameters for each classifier. The optimal hyperparameter configurations discovered through this process are summarized in Table~\ref{tab:hyperparam_search}.


\begin{longtable}[c]{llll}
\caption{Hyperparameter table}\\
\hline
Model & Hyperparameter & Search space & Best value\\
\hline
\hline
\endfirsthead

\hline
Model & Hyperparameter & Search space & Best value\\
\hline
\hline
\endhead

\hline
\endfoot

\endlastfoot

\multirow{3}{*}{Logistic Regression} 
 & C               & 0.1, 1, 10, 100                        & 100 \\
 & penalty         & l1, l2                             & l2 \\
 & class weight    & None, balanced                       & None \\
\hline

\multirow{4}{*}{Decision Tree} 
 & max\_depth         & None, 2, 5, 10, 20                     & 2 \\
 & min\_samples\_split & 2, 5, 10                              & 10 \\
 & criterion          & gini, entropy, log\_loss         & entropy \\
 & class weight       & None, balanced                     & balanced \\
\hline

\multirow{5}{*}{Random Forest} 
 & n\_estimators       & 50, 100, 150                          & 100 \\
 & max\_depth          & None, 2, 5, 10, 20                     & 10 \\
 & min\_samples\_split & 2, 5, 10                              & 2 \\
 & criterion           & gini, entropy, log\_loss         & entropy \\
 & class\_weight       & None, balanced                      & balanced \\
\hline

\multirow{4}{*}{SVC} 
 & C              & 0.1, 1, 10                             & 10 \\
 & kernel         & linear, poly, rbf, sigmoid     & rbf \\
 & gamma          & scale, auto                        & scale \\
 & class\_weight  & None, balanced                       & None \\
\hline

\multirow{3}{*}{Gradient Boosting} 
 & n\_estimators  & 50, 100, 150                          & 100 \\
 & learning\_rate & 0.001, 0.01, 0.1                      & 0.1 \\
 & max\_depth     & 3, 5                                  & 3 \\
\hline

\multirow{5}{*}{Multi-Layer Perceptron} 
 & hidden\_layer\_sizes & (8,), (16,), (8,16), (16,32)          & (16, 32) \\
 & activation           & relu, tanh                       & relu \\
 & learning\_rate\_init & 0.001, 0.01, 0.1                     & 0.1 \\
 & alpha                & 0.0001, 0.001                        & 0.0001 \\
 & solver               & sgd, adam                       & adam \\
\hline

\multirow{7}{*}{XGBoost} 
 & n\_estimators        & 50, 100, 150                         & 50 \\
 & learning\_rate       & 0.001, 0.01, 0.1, 0.5               & 0.5 \\
 & max\_depth           & 2, 3, 5, 7                          & 2 \\
 & subsample            & 0.8, 1.0                            & 0.8 \\
 & colsample\_bytree    & 0.8, 1.0                            & 1 \\
 & reg\_lambda          & 0, 1                                & 0 \\
 & min\_child\_weight   & 1, 3, 5                            & 3 \\
\hline

\multirow{5}{*}{LigthGBM} 
 & n\_estimators    & 50, 100, 150                         & 150 \\
 & learning\_rate   & 0.01, 0.1                            & 0.1 \\
 & max\_depth       & 2, 3, 5                            & 3 \\
 & num\_leaves      & 5, 10, 15, 20                      & 10 \\
 & subsample        & 0.8, 1.0                           & 0.8 \\
\hline

\multirow{3}{*}{Adaptive Boosting} 
 & n\_estimators     & 50, 100, 200                         & 100 \\
 & learning\_rate    & 0.001, 0.01, 0.1, 0.5, 1.0          & 0.01 \\
 & estimator         & \makecell[t l]{Decision Tree, \\ Random Forest} & Random Forest \\
\hline
\label{tab:hyperparam_search}
\end{longtable}

\begin{table}[!ht]
\centering
\caption{Feature transformation results}
\label{tab:grade_feature_extractors_count_PCA}
\begin{tabular}{@{}lcc@{}}
\hline
\multirow{2}{*}{Feature Extractor} & \multicolumn{2}{c}{Feature count}\\ \cmidrule(l){2-3}
& Extracted & PCA \\
\hline\hline
InceptionResNetV2          & 1536   & 125             \\
Xception                   & 2048   & 156      \\
ResNet101                  & 2048   & 129           \\
ResNet152                  & 2048   & 131\\
MobileNetV2                & 1280   & 163           \\
DenseNet169                & 1664   & 166  \\
InceptionV3                & 2048   & 160           \\
ResNet152V2                & 2048   & 74  \\
EfficientNetB7             & 2560   & 104           \\
ConvNeXtTiny               & 768    & 138          \\
ConvNeXtLarge              & 1536   & 132           \\
DenseNet201                & 1920   & 157           \\
\hline
\end{tabular}
\end{table}
Table \ref{tab:grade_feature_extractors_count_PCA} shows the number of features generated by different feature extractors. However, the raw feature dimensions are quite large, which can make model training computationally expensive and less efficient. To address this, we applied Principal Component Analysis (PCA) to reduce the dimensionality of the features. This reduction helps create a more efficient training while preserving the most important information from the original features.
\subsection{Selection of Feature Extractors}
The effectiveness of pre-trained convolutional neural networks (CNNs) for feature extraction varies significantly due to architectural differences, which influence how well models capture either object-specific or domain-specific features~\citep{heckler2023exploring, guerin2021combining}. To identify the most suitable feature extractors for egg quality classification, we evaluated a range of state-of-the-art pre-trained models on their ability to generate informative representations from egg images.
Each model was used to extract features from the image dataset, followed by dimensionality reduction using PCA, and then classified using a Random Forest classifier. Random Forest was selected for this comparison due to its consistent and strong performance on PCA-transformed feature spaces within our preliminary experiments.

The comparative performance of the pre-trained models in the context of egg grade classification is presented in Table~\ref{tab:grade_feature_extractors_comparison}. This analysis guided the selection of the most effective image-based feature extractor for inclusion in the final ensemble framework.

\begin{table}[!ht]
\centering
\caption{Performance comparison of pre-trained CNN models as feature extractors for egg grade classification using the Random Forest classifier.}
\label{tab:grade_feature_extractors_comparison}
\begin{tabular}{@{}lc@{}}
\hline
Feature Extractor & Accuracy (\%)   \\ 
\hline\hline
InceptionResNetV2          & 75.00                \\
Xception                   & 72.22              \\
ResNet101                  & 72.69              \\
ResNet152                  & \textbf{79.63}     \\
MobileNetV2                & 75.46              \\
DenseNet169                & \textbf{79.17}     \\
InceptionV3                & 72.22              \\
ResNet152V2                & \textbf{77.31}     \\
EfficientNetB7             & 76.39              \\
ConvNeXtTiny               & 73.15              \\
ConvNeXtLarge              & 71.76              \\
DenseNet201                & 75.93              \\
\hline
\end{tabular}
\end{table}

The results indicate that ResNet152 achieved the highest accuracy of 79.63\%, followed closely by DenseNet169 at 79.17\%, and ResNet152V2 at 77.31\%. These top three models significantly outperformed others, such as ConvNeXtLarge (71.76\%) and Xception (72.22\%), with ResNet152’s accuracy being approximately 4\% higher than EfficientNetB7 (76.39\%) and 4.63\% higher than InceptionResNetV2 (75.00\%). The strong performance of ResNet-based models (ResNet152 and ResNet152V2) suggests that the residual learning framework may be particularly effective for this task. The range of accuracies across all models, from 71.76\% to 79.63\%, highlights the variability in feature extraction effectiveness. 

Based on their superior performance, \textbf{ResNet152}, \textbf{DenseNet169}, and \textbf{ResNet152V2} were selected as the final feature extractors for the framework. Feature vectors from each model were independently extracted for every image, and their prediction outputs were subsequently integrated during the ensemble stage to determine the final classification result.

\subsection{Egg Grading Classification - Tabular vs Image vs Multimodal}
To evaluate the effectiveness of different data modalities in predicting egg grading, we conducted comparative experiments using three distinct input types: (i) tabular features (i.e., egg shape and weight), (ii) image-based features, and (iii) a multimodal fusion of both image and tabular features presented in Table~\ref{tab:grade-classification-comparision}.
For both the image-based and multimodal approaches, features were extracted using the top-performing pre-trained CNN models identified in earlier experiments: ResNet152, DenseNet169, and ResNet152V2. These models were selected based on their superior performance in extracting discriminative visual representations relevant to egg quality.

The image-based approach leverages deep feature embeddings derived from these pre-trained CNN models to capture external egg characteristics such as shell color, texture, and contour. In contrast, the tabular approach relies solely on numerical descriptors of structural properties specifically, egg shape and weight. The multimodal framework integrates both visual and structural features through concatenation, enabling the model to learn richer and more holistic representations. 

\begin{table}[!ht]
\centering
\caption{Classification accuracy (\%) comparison across different input modalities (Tabular, Image only, and Multimodal) and classifiers for egg grading. Image-based and multimodal approaches used features extracted from ResNet152, DenseNet169, and ResNet152V2. The highest accuracy for each modality is highlighted in bold.}
\label{tab:grade-classification-comparision}
\resizebox{\textwidth}{!}{
\begin{tabular}{@{}lcccc ccc@{}}
\toprule
\multirow{2}{*}{Classifier} & \multirow{2}{*}{\makecell{Tabular}}  & \multicolumn{3}{c}{Image only}  & \multicolumn{3}{c}{Multimodal} \\ 
\cmidrule(l){3-5} \cmidrule(l){6-8} 
& & ResNet152 & DenseNet169 & ResNet152V2 & ResNet152 & DenseNet169 & ResNet152V2\\
\hline \hline
Logistic Regression  &           62.96                     & 70.83             & 78.70                & 72.69              & 68.98             & 77.31               & 71.30               \\
Decision Tree     &         62.04                & 67.59             & 57.87               & 62.96              & 75.46             & 66.67              & 62.04               \\
Random Forest    &          62.50                & \textbf{79.63}    & 79.17              & \textbf{77.31}     & 77.78             & 76.85              & 70.83               \\
SVC   &             62.50                                 & 75.00             & 78.70              & 71.76              & 75.00             & 79.17              & \textbf{75.93}      \\
Gradient Boosting &       60.65                  & 72.69             & 67.59              & 72.69              & 76.39             & 68.52              & 70.83               \\
Multi-Layer Perceptron &       65.74                       & 73.61             & \textbf{79.63}     & 74.54              & 71.76             & \textbf{81.48}     & 75.00               \\
XGBoost &          \textbf{66.67}                            & 77.31             & 73.15              & 74.54              & \textbf{80.56}    & 74.07              & 71.30               \\
LightGBM &           65.74                   & 76.85             & 67.13              & 71.76              & 73.61             & 71.30              & 68.52               \\
Adaptive Boosting &       61.11                           & 74.07             & 75.93              & 75.93              & 75.46             & 77.31              & 74.07               \\ \hline
\end{tabular}
}
\end{table}
 In the tabular-only approach, XGBoost came out on top with an accuracy of 66.67\%. Its performance was just slightly better than Multi-Layer Perceptron and LightGBM, which both scored 65.74\%. This shows that while all three models performed fairly close to each other, XGBoost had a small edge when working only with tabular features.
 
 Within the image-based approach, Random Forest delivered the highest accuracies for both ResNet152 (79.63\%) and ResNet152V2 (77.31\%), respectively. For DenseNet169, the Multi-Layer Perceptron (MLP) model excelled, also reaching 79.63\% accuracy, outperforming other models such as Decision Tree (57.87\%), which exhibited lower performance across all feature extractors. The range of accuracies for DenseNet169 (57.87\% to 79.63\%) highlights the importance of selecting an appropriate prediction model. Consequently, Random Forest with ResNet152 and ResNet152V2, and MLP with DenseNet169, were chosen for the final training and the aggregation of results in the image-based approach. 

The integration of tabular features with images (multimodality) significantly enhanced prediction performance. Notably, the accuracy of the MLP model with DenseNet169 improved from 79.63\% in the image-based approach to 81.48\% in the multimodal setting, a 1.85\% increase. Similarly, XGBoost with ResNet152 saw around 1\% improvement, reaching 80.56\% accuracy in multimodal approach from 79.63\% in the image-based approach. However, the image feature set from ResNet152V2 combined with tabular features underperformed compared to the Support Vector Classifier (SVC) achieving 75.93\% in contrast to 77.31\% in the image-based approach. Compared to the image-based approach's performance, Random Forest model struggled to properly generalize on multimodal input, potentially due to its sensitivity to the heterogeneous patterns between image-extracted and tabular features. 

Experimental results reveal that the multimodal approach consistently outperformed both unimodal variants. While the image-based models demonstrated strong performance individually, the inclusion of tabular features further enhanced predictive accuracy by providing complementary information. This finding underscores the effectiveness of fusing heterogeneous data sources to improve classification performance in complex real-world tasks.
\subsection{Egg Grading with Ensemble Approach}

Finalizing the result for image-only and multimodal approaches, we applied majority voting to combine the top three prediction results, as detailed in Table~\ref{tab:grade_final_ensemble_result}. For the image-based approach, the ensemble of Random Forest with ResNet152 and ResNet152V2, and MLP with DenseNet169, achieved an accuracy of 85.19\%. 
\begin{table}[!b]
\centering
\caption{Comparative Egg Grade Performance of Ensemble Approach: Image-based vs. Multimodal}
\label{tab:grade_final_ensemble_result}
\begin{tabular}{@{}llllc@{}}
\hline
Approach         & Data Modality                & Feature Extractor & \multicolumn{1}{l}{Best Classifier} & Accuracy (\%)       \\
\hline \hline
\multirow{6}{*}{Ensemble} & \multirow{3}{*}{Image only}      & ResNet152                  & Random Forest                       & \multirow{3}{*}{85.19} \\
                          &                                  & DenseNet169                & Multi-Layer Perceptron                                 &                         \\
                          &                                  & ResNet152V2                & Random Forest                       &                         \\ \cmidrule(l){2-5} 
                          & \multirow{3}{*}{\makecell[c l]{Multimodal\\(image + weight and shape)}} & ResNet152                  & XGBoost                                 & \multirow{3}{*}{\textbf{86.57}} \\
                          &                                  & DenseNet169                & Multi-Layer Perceptron                                 &                         \\
                          &                                  & ResNet152V2                & SVC                                           &                         \\ \bottomrule
\end{tabular}
\end{table}
For the multimodal approach, the ensemble of XGBoost with ResNet152, MLP with DenseNet169, and SVC with ResNet152V2 yielded a superior accuracy of 86.57\%, a 5.09\% improvement over the best individual multimodal model (MLP with DenseNet169 at 81.48\%). These results highlight the effectiveness of the multimodal ensemble approach in leveraging complementary strengths of image and tabular features to achieve superior performance. 

\begin{figure}[!ht]
\centering
\begin{subfigure}{0.48\textwidth}
    \includegraphics[width=\textwidth]{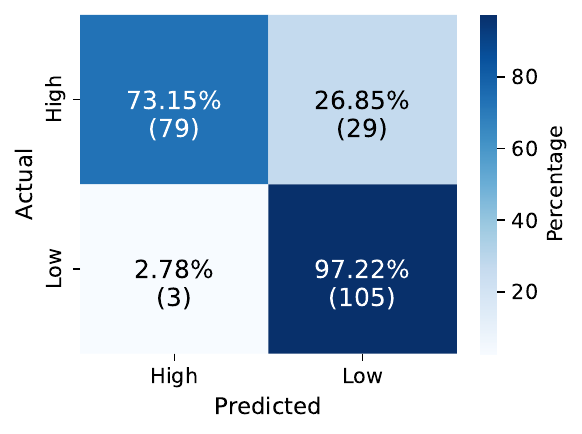}
    \caption{Image-based ensemble result}
    \label{fig:grade_image_based_ensemble_conf}
\end{subfigure}
\hfill
\begin{subfigure}{0.48\textwidth}
    \includegraphics[width=\textwidth]{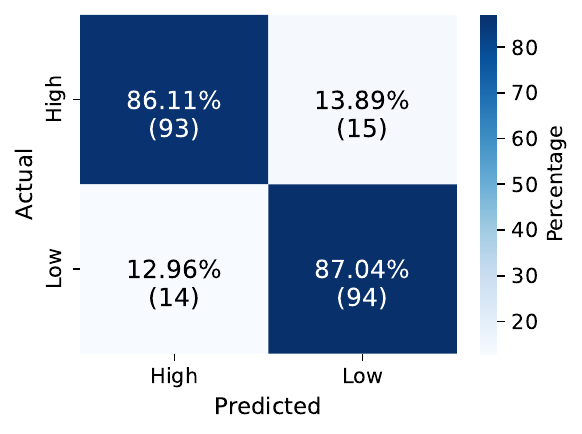}
    \caption{Multimodal ensemble result}
    \label{fig:grade_multimodal_ensemble_conf}
\end{subfigure}
\caption{Confusion matrix of egg grade classification ensemble result for different data modality }
\label{fig:grade_ensemble_confs}
\end{figure}

The Figure~\ref{fig:grade_ensemble_confs} demonstrates the confusion matrices of the ensemble result of the models described in Table~\ref{tab:grade_final_ensemble_result} in image-based and multimodal setting. For image input data modality in Figure~\ref{fig:grade_image_based_ensemble_conf}, the True Negative rate has increased significantly from 72.22\% in tabular approach to 97.22\%, a 25\% increase. Besides, the False Positive rate has decreased by 25\% and came down to 2.78\%. Overall, the ensemble approach in image-based model has improve the prediction result notable but become slightly biased towards ``Low" class.

The confusion matrix for the multimodal input-based ensemble approach, described in Figure~\ref{fig:grade_multimodal_ensemble_conf}, demonstrates a well-balanced performance with a True Positive rate of 86.11\% for the ``High" class and a True Negative rate of 87.04\% for the ``Low" class. 

\begin{figure}[!ht]
    \centering
    \includegraphics[width=0.6\textwidth]{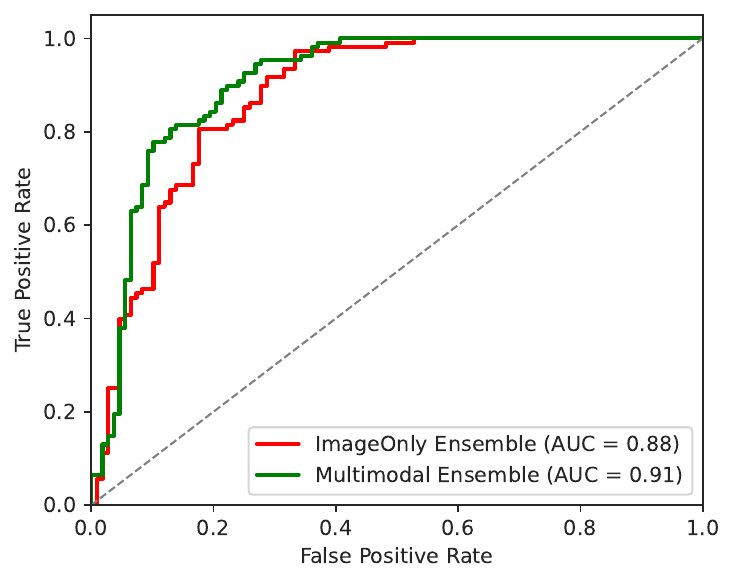}
    \caption{Combined ROC-AUC curve for Egg grade classification}
    \label{fig:grade_roc_curve}
\end{figure}

Furthermore, the low False Positive rate (12.96\%) and False Negative rate (13.89\%) reflect minimal misclassification errors, indicating robust generalization across both target classes. The ROC-AUC curve in Figure~\ref{fig:grade_roc_curve}, created using the average predicted probabilities, further confirms the superiority of the multimodal ensemble model compared to other approaches, highlighting its effectiveness in egg grading.

The figure compares the classification performance of two ensemble models (Image-only ensemble and Multimodal ensemble) for egg grading. Both ROC curves lie well above the diagonal reference line, indicating that the models perform significantly better than random guessing. The Image-only ensemble achieves an AUC of 0.88, reflecting strong discriminative ability based on visual features alone. Importantly, the Multimodal ensemble further improves the AUC to 0.91, demonstrating the added value of integrating tabular features (weight and shape) alongside image-derived features. This consistent upward shift in the ROC curve highlights better sensitivity–specificity trade-offs, particularly at lower false-positive rates, thereby providing strong evidence of robust generalization and suggesting that the multimodal approach achieves a more effective balance between sensitivity and specificity.

\subsection{Egg Freshness Classification - Tabular vs Image vs Multimodal}
This subsection presents the results of the egg freshness classification based on the Yolk Index (YI), a critical indicator of internal egg quality, calculated as the ratio of yolk height to diameter. The dataset for this task comprised of 192 samples, with balanced representation across egg freshness categories, as detailed in Section~\ref{preprocessing_augmentation}. Consistent with the Haugh Unit (HU)-based classification task, multiple machine learning models, including Random Forest, Multi-Layer Perceptron (MLP), XGBoost, and others, were evaluated using 10-fold cross-validation. The analysis compares the effectiveness of individual input modalities (image-based and tabular data-based) and multimodal inputs in predicting egg freshness. These findings are significant for quality control in the poultry industry, where accurate freshness assessment ensures consumer safety and product reliability.

To evaluate the impact of different input modalities on predicting egg freshness, we conducted experiments using tabular features, image-based features, and multimodal fusion, with results presented in Table~\ref{tab:freshness-classification-comparision}. The image-based and multimodal approaches utilized features extracted from ResNet152, DenseNet169, and ResNet152V2, which had demonstrated strong representation capability in earlier experiments (Table \ref{tab:grade-classification-comparision}).   

\begin{table}[!ht]
\centering
\caption{Classification accuracy (\%) comparison across different input modalities (Tabular, Image only, and Multimodal) and classifiers for egg freshness. Image-based and multimodal approaches used features extracted from ResNet152, DenseNet169, and ResNet152V2. The highest accuracy for each modality is highlighted in bold.}
\label{tab:freshness-classification-comparision}
\resizebox{\textwidth}{!}{
\begin{tabular}{@{}lcccc ccc@{}}
\toprule
\multirow{2}{*}{Classifier} & \multirow{2}{*}{\makecell{Tabular}}  & \multicolumn{3}{c}{Image only}  & \multicolumn{3}{c}{Multimodal} \\ 
\cmidrule(l){3-5} \cmidrule(l){6-8} 
& & ResNet152 & DenseNet169 & ResNet152V2 & ResNet152 & DenseNet169 & ResNet152V2\\
\hline \hline
Logistic Regression  &    60.94                            & 60.94             & \textbf{67.71}                & 66.67              & 67.71             & 68.75               & 59.38               \\
Decision Tree     &     59.38                    & 55.73             & 57.29               & 57.89              & 59.90             & 56.25              & 55.73               \\
Random Forest    &     56.77                     & 60.94    & 62.50              & \textbf{67.71}     & 65.62             & 61.46              & 57.29               \\
SVC   &       60.94                                       & 61.46             & 61.98              & 64.06              & 66.67             & \textbf{70.31}              & 63.54      \\
Gradient Boosting &    60.42                     & 56.77             & 57.81              & 58.33              & 65.62             & 55.73              & 55.21               \\
Multi-Layer Perceptron &     60.94                         & \textbf{66.15}             & 63.54     & 67.19              & 65.62             & 69.27     & \textbf{63.54}               \\
XGBoost &       61.98                               & 64.58             & 64.58              & 66.15              & \textbf{67.19}    & 61.98              & 60.42               \\
LightGBM &          \textbf{64.06}                    & 59.38             & 61.98              & 60.94              & 62.50             & 58.33              & 53.65               \\
Adaptive Boosting &       57.81                           & 60.94             & 60.94              & 63.02              & 65.10             & 58.85              & 57.81               \\ \hline
\end{tabular}
}
\end{table}



In the tabular-only setting, classification accuracy remained modest, with LightGBM achieving the highest score of 64.06\%, followed closely by XGBoost (61.98\%) and Logistic Regression/MLP (60.94\%). These results highlight the limitations of relying solely on structural traits such as shape and weight, which, while informative, fail to capture the nuanced external characteristics that distinguish freshness levels.

The image-based models provided notable performance improvements. Using features extracted from DenseNet169, Logistic Regression reached the highest overall accuracy of 67.71\%. Similarly, Random Forest performed best with ResNet152V2 features (67.71\%), and MLP delivered competitive performance across architectures, peaking at 67.19\% with ResNet152V2. These outcomes emphasize the discriminative power of visual features for freshness detection, though performance varied substantially depending on the choice of classifier and backbone CNN.

The multimodal fusion of image and tabular features consistently improved results compared to either unimodal approach. For instance, SVC with DenseNet169 features achieved 70.31\%, the highest accuracy across all modalities, marking a clear improvement over its image-only counterpart (61.98\%). Likewise, MLP with DenseNet169 improved from 63.54\% (image-only) to 69.27\%, while XGBoost with ResNet152 increased from 64.58\% to 67.19\%. Notably, some classifiers such as Random Forest and Gradient Boosting showed less benefit in the multimodal setting, suggesting potential difficulties in handling heterogeneous feature types. However, the overall accuracies were modest, with MLP and ResNet152V2 in the multimodal approach achieving the lowest accuracy of 63.54\%, possibly due to challenges in integrating heterogeneous image and tabular features. Notably, DenseNet169 demonstrated robust performance across both image-only and multimodal modalities, underscoring its effectiveness as a feature extractor for egg freshness prediction. 

These results underscore the superior effectiveness of a multimodal approach, likely due to their ability to capture visual representations by combining with physical features relevant to freshness, which has practical implications for automated quality assessment in the poultry industry.

\subsubsection{Egg Freshness Assessment with Ensemble Approach}  

To enhance prediction performance, majority voting was applied to combine the top three prediction results from image-only and multimodal approaches, as shown in Table~\ref{tab:freshness_Final_Ensemble_Result}. The image-based ensemble, comprising MLP with ResNet152, Logistic Regression with DenseNet169, and Random Forest with ResNet152V2, achieved an accuracy of 67.71\%, showing no improvement over the best individual model. The multimodal ensemble, including XGBoost with ResNet152, SVC with DenseNet169, and MLP with ResNet152V2, reached a slightly higher accuracy of 70.83\%, a marginal 0.52\% increase over the best individual multimodal model.  
\begin{table}[!ht]
\centering
\caption{Comparative Egg Freshness Performance of Ensemble Approach: Image-Only vs. Multimodal}
\label{tab:freshness_Final_Ensemble_Result}
\begin{tabular}{llllc}
\hline
Approach         & Data Modality              & Feature Extractor & \multicolumn{1}{l}{Best Prediction Model} & Accuracy (\%)      \\ 
\hline \hline
\multirow{6}{*}{Ensemble} & \multirow{3}{*}{Image only}      & ResNet152                  & Multi-Layer Perceptron                                 & \multirow{3}{*}{67.71} \\
                          &                                  & DenseNet169                & Logistic Regression                            &                         \\
                          &                                  & ResNet152V2                & Random Forest                       &                         \\ \cline{2-5} 
                          & \multirow{3}{*}{\makecell[c l]{Multimodal \\(image + weight and shape index)}} & ResNet152                  & XGBoost                                 & \multirow{3}{*}{\textbf{70.83}} \\
                          &                                  & DenseNet169                & SVC                                           &                         \\
                          &                                  & ResNet152V2                & Multi-Layer Perceptron                                 &                         \\ \hline
\end{tabular}
\end{table}

\begin{figure}[!b]
\centering
\begin{subfigure}{0.48\textwidth}
    \includegraphics[width=\textwidth]{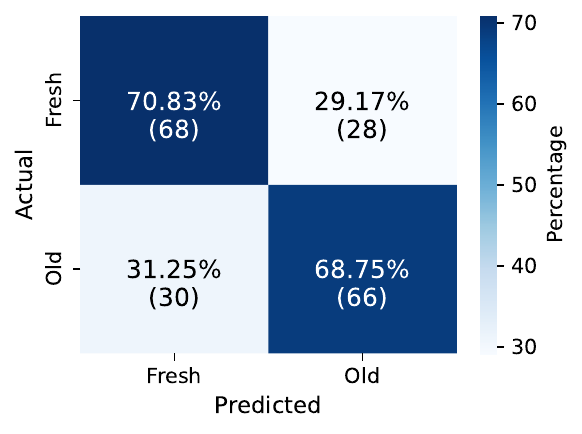}
    \caption{Image-based ensemble result}
    \label{fig:image_based_ensemble_conf}
\end{subfigure}
\hfill
\begin{subfigure}{0.48\textwidth}
    \includegraphics[width=\textwidth]{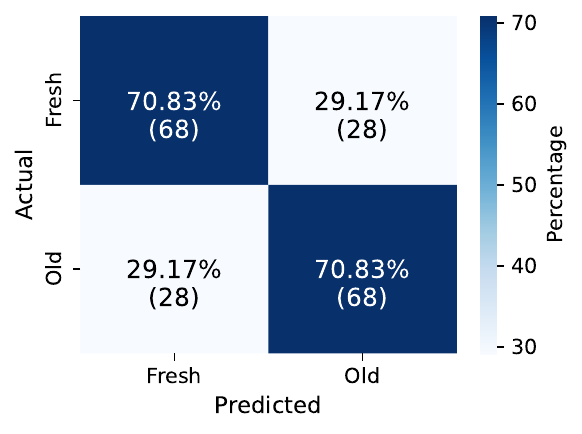}
    \caption{Multimodal ensemble result}
    \label{fig:multimodal_ensemble_conf}
\end{subfigure}
\caption{Confusion matrix of egg freshness classification ensemble result for different data modality }
\label{fig:Freshness_ensemble_confs}
\end{figure}

\begin{figure}[t]
    \centering
    \includegraphics[width=0.6\textwidth]{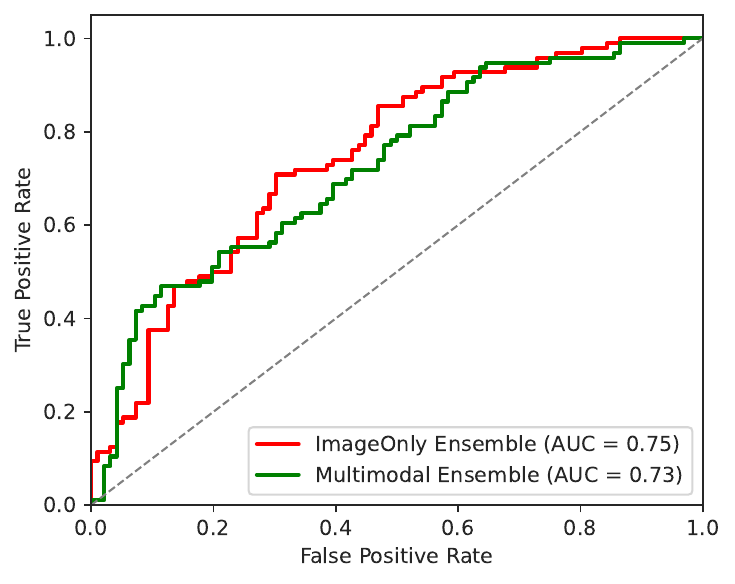}
    \caption{Combined ROC-AUC curve for Egg freshness classification}
    \label{fig:freshness_roc_curve}
\end{figure}

The confusion matrices in Figure~\ref{fig:Freshness_ensemble_confs} and the ROC curve in Figure~\ref{fig:freshness_roc_curve} reveal distinct performance trends across the three egg freshness classification approaches. The AUC score has been calculated by taking the average of the three prediction probabilities. The image-based ensemble achieved the highest discriminative ability with an AUC of 0.75, outperforming the multimodal ensemble (AUC = 0.73), suggesting that visual features can also capture freshness cues robustly without any extra information. However, the multimodal ensemble demonstrated improved classification for ``Old" eggs, correctly identifying 70.83\% (68 instances) compared to the image-based model’s 68.75\% (66 instances). In both approaches, the number of fresh eggs that were also identified as fresh was the same, which is 70.83\%. The multimodal approach exhibited marginally better balance in precision-recall trade-offs. 
High accuracy of multimodal ensemble indicates that the model's final prediction is often correct. However, the low AUC reveals that the model's probability scores are not well-separated among the classes. This means the model frequently assigns high probabilities to one class for instances that actually belong to other classes, hurting its ranking ability. Hence, the multimodal ensemble model is effective at making the final ``hard" decision but poor at quantifying its certainty.

The limited improvement in ensemble performance suggests potential overlap in the predictive information provided by the models and insufficient diversity in their outputs. These modest results may reflect challenges in capturing subtle freshness variations based on the Yolk Index, highlighting the need for further exploration of alternative feature extractors, additional tabular features, or more sophisticated ensemble techniques to improve classification performance in future work.

\section{Discussion}
\subsection{Findings and Observations}
This study presents several novel and significant findings that advance the understanding of egg freshness and grading using computer vision and multimodal learning. Most notably, our study provides the first systematic evidence that the external RGB appearance of an intact egg encodes reliable information about both its grade (interior quality) and freshness (temporal deterioration). We demonstrate that computer-vision pipelines can extract subtle spectral and textural signatures that correlate with Haugh Unit and Yolk Index values, achieving 86.5\% and 70.83\%. This finding opens an entirely non-invasive avenue for quality assessment that bypasses the need for candling or destructive testing as presented in previous researches~\citep{harnsoongnoen2021grades, nematinia2018assessment, tolentino2018development, vasileva2018assessing, omid2013expert, ab2018automated, tang2024non}.

A key observation is that traditional tabular features, weight and shape index, long-standing tabular benchmarks in egg-grading standards~\citep{ab2018automated, abaci2023examination, de2023non}, are insufficient for reliable classification, as demonstrated by their poor predictive performance in this study. In contrast, image-based features significantly improve model generalizability, suggesting that visual cues play a dominant role in determining egg quality. This aligns with recent trends in food quality assessment, where deep learning has uncovered non-intuitive visual biomarkers for perishable goods~\citep{yang2023lightweight, shu2025fruit, shen2025advanced, shehzad2025computer},

Interestingly, our results challenge the conventional approach of directly using pre-trained model embeddings for prediction~\citep{rahman2021deep, poola2023covid}. Directly feeding ImageNet embeddings into a shallow classifier yielded mediocre results, whereas dimensionality reduction through PCA compression improved performance and appears to suppress nuisance variance such as illumination shifts or camera white-balance artifacts, while preserving the low-rank manifold on which egg-quality variations reside. This implies that while deep learning models capture rich representations, strategic feature transformation is critical to eliminate redundancy and amplify task-relevant signals. This observation also aligns with recent findings in few-shot agricultural vision tasks, where compact latent spaces often outperform high-dimensional raw features~\citep{ouyang2025few, zhou2025novel, zeng2025few}.

Furthermore, not all pre-trained architectures perform equally as feature extractors ~\citep{holliday2020pre}. Our experiments reveal that ResNet-based models (ResNet152 and ResNet152V2) consistently outperform alternatives (e.g., InceptionV3, VGG19, EfficientNet) in this task. Residual skip connections evidently facilitate the capture of global contextual features and fine-grained object-specific details~\citep{xu2024development, shafiq2022deep, he2016deep}, a dual requirement that depthwise-separable or heavily down-sampled networks struggle to satisfy in this task.

A central objective of this study is to evaluate multimodal learning for egg classification. Our results demonstrate that combining image features with physical measurements (weight and shape index) yields better performance than unimodal inputs, underscoring the complementary nature of visual and tabular data. Thus, multimodal integration functions as a reliability-weighted ensemble, where the image data dominates when visual patterns are salient and the tabular data rescues edge cases that violate chromatic priors~\citep{hager2023best}.

Surprisingly, we observed that egg grade classification (haugh unit-based) exhibits a stronger correlation with external imagery than freshness classification (yolk index-based), suggesting that yolk-index-related degradation is more challenging to identify than haugh-unit-related changes. This discrepancy may arise because grade-related traits manifest more visibly in shell appearance, whereas freshness indicators require deeper sensory analysis~\citep{gao2025research}. This finding has practical implications for prioritizing computer vision in grading over freshness assessment.

Finally, our analysis of egg source characteristics revealed a significant correlation between procurement channels and quality metrics. Specifically, eggs sourced from supermarkets and grocery stores exhibited consistently lower grades and freshness levels compared to those obtained from wholesale markets. This trend suggests that prolonged storage even under controlled environmental conditions leads to measurable degradation in perishable quality over time~\citep{aschemann2018consumer}. These findings highlight a critical gap in current retail inventory management practices, where extended shelf-life appears to compromise product quality~\citep{vandevijvere2023unhealthy}. From a practical standpoint, this discovery underscores the need for supermarkets and grocery retailers to implement more frequent restocking cycles. Such operational adjustments would not only improve the freshness of eggs available to consumers but also align with growing market demands for higher-quality perishable goods. This recommendation carries particular weight, given increasing consumer awareness about food quality and the demonstrated ability of our models to detect subtle quality differences through visual inspection.

\subsection{Limitations and Future Directions}
Although this work introduces a new path of research for the poultry industry with egg grade and freshness classification with visual and multimodal features, there are some limitations of this work that should be acknowledged. Additionally, a few suggestions on future research directions in this domain could help fellow researchers to contribute meaningfully to similar tasks.    

First, the dataset used in this research work is relatively smaller compared to other computer vision-based research works. Although the dataset encompasses egg collection from a wide variety of sources, the amount of data that could be ultimately used was limited. This occurred due to a lack of manpower and resources for storing and handling the eggs. Additionally, some eggs got broken in transportation and during the experiment. Therefore, we had to opt for an augmentation technique to increase the amount of data. A large dataset could help the models to better generalize for the task and improve prediction performance, opening the door for industry-grade applications.

Second, our team collected eggs from different sources without knowing the exact time difference between egg collection and the time the chicken laid those eggs. Therefore, it was found that the dataset contained more eggs of ``Low" grade and ``Old' freshness class than its counterpart.     

Third, from our experimental results, it becomes evident that egg external RGB images carry more relational visual features with grade than with freshness. In this work, we had assumed that the visual feature significance could be of equal importance. Consequently, we only evaluated the usability of pre-trained CNNs for the grading task and used their best-performing feature extractors directly for freshness tasks. An in-depth experiment for the freshness assessment could reveal more interesting facts about egg freshness classification using external RGB images.     

Fourth, in this study, we used pre-trained CNN models exclusively for visual feature extraction rather than conducting end-to-end training on RGB images. This design choice was necessitated by the limited size of our dataset, as fine-tuning deep CNN architectures typically requires substantially larger training sets to achieve robust generalization. While transfer learning through pre-trained models provided effective feature representations, we acknowledge that more advanced approaches such as vision transformers (ViTs)~\citep{dosovitskiy2020image}, few-shot learning~\citep{parnami2022learning}, or zero-shot learning paradigms~\citep{xian2017zero}, could potentially yield superior performance given adequate training data. These emerging techniques may offer enhanced capability in capturing subtle discriminative features for egg quality classification, particularly in scenarios where dataset constraints can be addressed through expanded collection or synthetic data augmentation. Future work should explore these directions to further advance the accuracy and generalizability of visual-based egg quality assessment systems.  

Fifth, we implemented feature fusion through the simple concatenation of image-derived and tabular features. While this approach provided a baseline for multimodal integration, we recognize that more sophisticated fusion techniques, such as element-wise multiplication, weighted addition, or learned fusion via convolutional layers, could potentially enhance the model's predictive performance. These alternative fusion strategies may better capture non-linear interactions between heterogeneous feature types, thereby improving the model's ability to learn complementary relationships between visual and physical egg characteristics. Future investigations should systematically evaluate these fusion methods to optimize information integration while maintaining computational efficiency.   

\section{Conclusion}
This study proposed ELMF4EggQ, an ensemble-based multimodal machine learning framework for non-destructive egg quality assessment, focusing on both egg grade and freshness classification. The framework integrates visual features from external egg images with structural attributes (shape and weight), leveraging their complementary strengths through feature-level fusion and ensemble learning.
A novel dataset of 186 brown-shelled eggs was developed for this purpose, with expert-labeled grade and freshness annotations derived from internal quality measurements, including yolk index and Haugh unit. This dataset, along with the full codebase, has been made publicly available to promote transparency, reproducibility, and further research in this domain.

Through extensive experiments, we evaluated and compared unimodal (image-only and tabular-only) and multimodal inputs, with and without ensemble learning. Among unimodal approaches, the image-only model without ensemble achieved 79.63\% accuracy for grade and 67.71\% for freshness classification. Incorporating both image and tabular features improved grade classification to 81.48\%, though freshness classification slightly decreased to 66.15\%.
The most significant performance gains were observed with ensemble learning. The image-based ensemble achieved 85.19\% (grade) and 67.71\% (freshness), while the multimodal ensemble further boosted results to 86.57\% and 70.83\%, respectively. These findings confirm the effectiveness of multimodal fusion and ensemble strategies in enhancing classification robustness and generalization. 
Importantly, the results also suggest that image-based features contribute more strongly to predictive performance than tabular features, particularly in grade classification. This may be attributed to the visual correlation between Haugh unit–based grading and external egg appearance—an association that is less pronounced in freshness classification.

The proposed framework demonstrates strong potential for real-world deployment in commercial egg processing facilities, retail quality assurance systems, and supply-chain optimization, enabling scalable, data-driven decision-making. Future work will explore larger and more diverse datasets, real-time inference on edge devices, and broader cross-market applicability.
\bibliographystyle{elsarticle-harv}
\bibliography{reference}

\end{document}